\documentclass{article}
\usepackage{spconf,amsmath,amsthm,amssymb,graphicx}
\usepackage{xcolor}
\usepackage{multirow}
\usepackage[normalem]{ulem}
\usepackage{subfigure}
\usepackage{booktabs}
\usepackage{adjustbox}
\usepackage{wrapfig,soul} 
\usepackage{enumitem}
\usepackage{hyperref}
\usepackage{mathtools}
\useunder{\uline}{\ul}{}

\newcommand{\conf}{\texttt{Conf}}
\newcommand{\lms}{\texttt{LMS}}
\newcommand{\gde}{\texttt{MA}}

\theoremstyle{definition}
\newtheorem{definition}{Definition}[section]

\title{On the Efficacy of Generalization Error Prediction Scoring Functions}
\name{Puja Trivedi$^{\star}$ \qquad Danai Koutra$^{\star}$ \qquad Jayaraman J. Thiagarajan$^{\dagger}$}

\address{$^{\star}$ University of Michigan \\
$^{\dagger}$Lawrence Livermore National Laboratory\thanks{This work was performed under the auspices of the U.S. Department of Energy by the Lawrence Livermore National Laboratory under Contract No. DE-AC52-07NA27344. Supported by the ASCR Co-Design project. It was also partially supported by the National Science Foundation under CAREER Grant No.~IIS 1845491. Code available: \href{https://github.com/pujacomputes/icassp23-gengap.git}{\texttt{https://github.com/pujacomputes/icassp23-gengap.git}}}}

\begin{document}
%
\maketitle
\begin{abstract}
Generalization error predictors (GEPs) aim to predict model performance on unseen distributions by deriving dataset-level error estimates from sample-level scores. However, GEPs often utilize disparate mechanisms (e.g., regressors, thresholding functions, calibration datasets, etc), to derive such error estimates, which can obfuscate the benefits of a particular scoring function. Therefore, in this work, we rigorously study the effectiveness of popular scoring functions (confidence, local manifold smoothness, model agreement), independent of mechanism choice. 
We find, absent complex mechanisms, that state-of-the-art confidence- and smoothness- based scores fail to outperform simple model-agreement scores when estimating error under distribution shifts and corruptions. Furthermore, on realistic settings where the training data has been compromised (e.g., label noise, measurement noise, undersampling), we find that model-agreement scores continue to perform well and that ensemble diversity is important for improving its performance. 
Finally, to better understand the limitations of scoring functions, we demonstrate that simplicity bias, or the propensity of deep neural networks to rely upon simple but brittle features, can adversely affect GEP performance.
Overall, our work carefully studies the effectiveness of popular scoring functions in realistic settings and helps to better understand their limitations. 
\end{abstract}
\section{Introduction}
\label{sec:intro}
Safe deployment of machine learning models requires suitable failure indicators so that models whose performance falls below an acceptable tolerance can be temporarily pulled from production.
While learning-theoretic complexity measures can be used to estimate model performance under \textit{i.i.d} assumptions~\cite{Jiang20_FantasticGen, Garg22_UnlabeldPred}, they are currently insufficient for estimating model generalization on \textit{out of distribution (o.o.d)} data.  
To this end, generalization error predictors (GEPs), which are designed to estimate performance on \textit{arbitrary} target datasets, have become popular~\cite{Ng22_LocalManifoldSmoothness,Guillory21_DoC,Chen21_Mandoline,Deng21_labelsnecessary,Djurisic22_ActivationShaping}.

\begin{figure}
    \centering
    \includegraphics[width = 0.85\columnwidth]{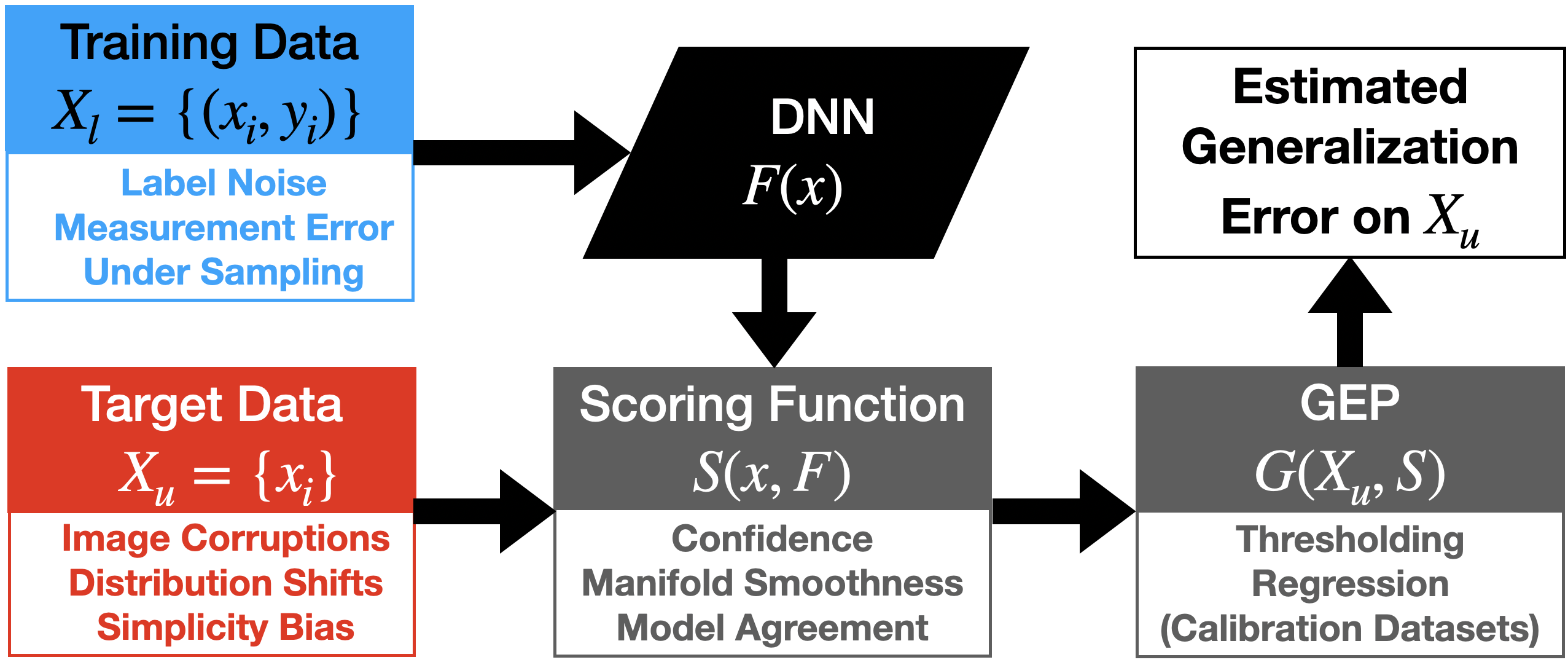}
    \caption{\textbf{Generalization Error Prediction.} We focus on the problem of generalization error prediction with classifiers and study the design of scoring functions under various distribution shifts, corruptions, and data fidelity issues.}
    \label{fig:overview}
    \vspace{-0.3cm}
\end{figure}

In brief, GEPs \textit{aggregate} sample-level \textit{scores} to predict generalization error on unlabeled target datasets (see Fig. \ref{fig:overview}).
Popular scoring functions, such as manifold proximity \cite{Deng21_labelsnecessary}, confidence estimates \cite{Guillory21_DoC}, local manifold smoothness \cite{Ng22_LocalManifoldSmoothness}, and agreement between independently trained models \cite{Jiang22_Disagreement,Chen21_SelfTrainEnsemb,Nakkiran20_DistGen}, attempt to measure the likelihood that the predicted label of a sample is correct. 
GEPs then use different mechanisms (thresholding functions, regressors, calibration datasets) to create dataset-level error estimates from the provided scores. 
However, since these mechanisms can vary in complexity, the efficacy of a particular scoring functions can obfuscated. 
For example, state-of-the-art GEPs often rely upon multiple, labeled calibrated datasets~\cite{Deng21_labelsnecessary,Deng21_RotPred,Guillory21_DoC,Lakshminarayanan17_DeeEns}, which can provide additional information that bolsters the performance of otherwise subpar scoring functions.
Therefore, in this paper, we use a simple, fixed GEP, and rigorously study the effectiveness of popular scoring functions (confidence, local manifold smoothness, model agreement) in several realistic settings and identify a potential cause of poor GEP performance.

\noindent \textbf{GEP performance under distribution shifts (Sec. \ref{sec:gep_dist}):} Using a large family of image-level corruptions and distribution shifts, we benchmark the three scoring functions and provide key insights on their efficacy in practice.

\noindent \textbf{Impact of training data fidelity on GEP performance (Sec. \ref{sec:corrupted_training}):} Scoring functions directly depend on data and model properties. Therefore, we consider common data fidelity issues (label noise, measurement errors and sampling discrepancies) to study what role data quality plays on the GEP performance. 

\noindent \textbf{Effect of simplicity bias on GEP performance (Sec. \ref{sec:simp_bias}):} Deep neural networks are susceptible to relying upon simple, spurious features \cite{Shah20_SimplicityBias} at the expensive of robust generalization. We study the impact of this behavior on the efficacy of different scoring functions \cite{Shah20_SimplicityBias,Trivedi22_Adaptation}.

\section{Preliminaries}

\label{sec:preliminaries}
We begin by formally introducing the problem setting and scoring functions. 
Let $\mathcal{X}_u = \{\bar{\mathrm{x}}_i\}$ be an unlabeled, target dataset and $\mathcal{X}_{\ell} = \{(\mathrm{x}_i, \mathrm{y}_i)\}$ be a labeled training dataset where $y_i$ is one of $C$ classes. Further, let $F:x\rightarrow [0,1]^C$ be a model (e.g., DNN) trained on $\mathcal{X}_{\ell}$ that outputs softmax probabilities over $C$ classes. 
GEPs utilize scores computed from model features on the target data distribution, which are expected to be correlated with dataset performance. We focus on popular sample-level scoring functions,  $\mathrm{S}(\mathrm{x}; \mathrm{F}) \rightarrow \mathbb{R}$, and define them below.

\begin{itemize}[leftmargin=*]
    \item \textbf{Confidence} (\conf) \cite{Guillory21_DoC}. We can directly obtain a sample-level score from $F$ by using the maximum softmax probability (e.g., the predicted class's confidence): 
    \begin{equation}
    \mathrm{S}(\mathrm{x};\mathrm{F}) = \max \mathrm{F(x)}.
    \end{equation}
    
    \item \textbf{Local Manifold Smoothness} (\lms) \cite{Ng22_LocalManifoldSmoothness}. Let $q(\mathrm{x}^{\prime}|\mathrm{x})$ be a local probability distribution over augmented samples, $\mathrm{x}^{\prime}$, that can be generated from a given natural sample, $\mathrm{x}$. Then, the LMS score is defined as $\mathrm{S}(\mathrm{x}) = \mathbb{E}_{\mathrm{x}^{\prime} \sim q(\mathrm{x}^{\prime}|\mathrm{x})} \quad \mathbb{I}[\mathrm{F}(\mathrm{x}^{\prime}) =\mathrm{F}(\mathrm{x})]$, where $\mathbb{I}$ is the indicator function.
     Note, in practice, the expectation over $q$ is approximated by sampling set of $k$ augmented samples $\mathrm{X}^{\prime} \coloneqq \{\mathrm{x_j}^{\prime} \sim q(\mathrm{x}^{\prime}|\mathrm{x})\}_{j=1}^{k}$ and we define $q$ using RandAug \cite{Cubuk20_RandAug}.  
    \begin{equation}
    \mathrm{S}(\mathrm{x};\mathrm{F}) \approx \frac{1}{k}\sum_{j=1}^k \quad \mathbb{I}[\mathrm{F}(\mathrm{x_j}^{\prime}) =\mathrm{F}(\mathrm{x})].
    \end{equation}
    
    \item \textbf{Model Agreement} (\gde) \cite{Jiang19_GenGap,Jiang22_Disagreement, Chen21_SelfTrainEnsemb}. Let $F_{0\dots r}$ be a set of $r$ independently trained models (e.g., models are trained using $r$ different seeds). WLOG, let $F_0$ be the base model for which the accuracy is estimated. Then the score can be computed as: 
    \begin{equation}
    \mathrm{S}(\mathrm{x};\mathrm{F}) = \frac{1}{r-1} \sum_{i=1}^r \mathbb{I}[\mathrm{F_0}(\mathrm{x}) =\mathrm{F}_r(\mathrm{x})]
    \label{eq:ma}
    \end{equation} 
\end{itemize}

Given these sample-level scores, GEPs then estimate the performance of deep neural networks (DNNs) under a wide-variety of distribution shifts and are defined as follows.
\begin{definition}
For an unlabeled dataset $\mathcal{X}_u$, a generalization error predictor $\mathrm{G}(\mathcal{X}_u; \mathrm{S})$ returns the estimated error of the pretrained classifier $\mathrm{F}$, based on the scores from $\mathrm{S}(\mathcal{X}_u; F)$.
\vspace{-5pt}
\end{definition}

State-of-the-art approaches propose to curate multiple (labeled) calibration datasets or train multiple models (with different hyper-parameters) in order to construct a well-calibrated GEP~\cite{Deng21_labelsnecessary,Deng21_RotPred,Guillory21_DoC,Lakshminarayanan17_DeeEns}. However, it can be difficult to obtain such calibration datasets in practice and training multiple models is expensive. Moreover, using such strategies can obfuscate the effectiveness of a given scoring function. Therefore, we focus on a simple, popular \textit{thresholding-based} GEP. This GEP simply aggregates thresholded sample-level scores to obtain a dataset-level estimate: $\frac{1}{|X|}\sum_i \mathbb{I}(\mathrm{S}(\bar{\mathrm{x}}_i; \mathrm{F}) > \tau)$, where the threshold hyperparameter $\tau$ is identified by regressing $\mathrm{G}$ to recover the true accuracy on a pre-defined, validation dataset. Given this fixed GEP, we are ready to assess the performance of different scoring functions in a fair setting.  

\section{A Closer Look at Scoring Functions}
\label{sec:method}
As discussed in Sec. \ref{sec:intro}, it is critical to disentangle the scoring function from the GEP mechanism to understand the former's effectiveness. Using a fixed thresholding-based GEP, we evaluate the ability of different scoring functions to accurately predict generalization over various distribution shifts (Sec. \ref{sec:gep_dist}) and under the realistic setting of training on low fidelity data (Sec. \ref{sec:corrupted_training}). 

\noindent \textit{Experimental Setup.} For all experiments, CIFAR10 is the source distribution on which we train ResNet-18 for 200 epochs with lr=0.05. STL10, CIFAR10.1 and CIFAR-10-C~\cite{Hendrycks19_CIFAR10C} are the target distributions, for which we estimate the generalization performance. CIFAR10.1 and STL10 represent near and far distribution shifts respectively. CIFAR-10-C contains samples generated from $15$ different naturalistic corruptions, such as ``fog" or ``blur", applied at five severity levels. Increased severity corresponds to increased shift from the training data. In all experiments, we report the mean absolute error (MAE) between the true target accuracy and the predicted target accuracy. The threshold, $\tau$, is determined by optimizing the thresholding function to minimize prediction error on the CIFAR10 validation dataset. Note that all results are averaged over $10$ seeds. We compute \gde~using a 10-member ensemble and use $10$ augmentations (RandAug~\cite{Cubuk20_RandAug}) to compute \lms.

\begin{table}[t]
\centering
\caption{\small{\textbf{Assessing Scoring Functions Under Distribution Shifts.}} We report the mean absolute error between the true and estimated accuracies obtained with different scoring functions. We average CIFAR-10-C corruptions by severity. The top two methods are \textbf{bolded} and 
\underline{underlined} respectively.}\label{table:clean_data}
\begin{adjustbox}{max width=\columnwidth}
\begin{tabular}{lrrrrrrr}
 \toprule
    $\mathrm{S}(\mathrm{x}; \mathrm{F})$ &
  \multicolumn{1}{l}{STL10} &
  \multicolumn{1}{l}{CIFAR10.1} &
  \multicolumn{1}{l}{Sev. 1} &
  \multicolumn{1}{l}{Sev. 2} &
  \multicolumn{1}{l}{Sev. 3} &
  \multicolumn{1}{l}{Sev. 4} &
  \multicolumn{1}{l}{Sev. 5} \\
 \midrule
\conf &  {\ul 0.1488} & {\ul 0.0117}    & {\ul 0.0246}           & {\ul 0.0430}           & {\ul 0.0476}           & {\ul 0.0596}           & {\ul 0.0661}           \\
\lms         & 0.3438       & 0.0961          & 0.0942                 & 0.1350                 & 0.1704                 & 0.2147                 & 0.2739                 \\
\gde         & \textbf{0.0420}       & \textbf{0.0054} & \textbf{0.0115}        & \textbf{0.0189}        & \textbf{0.0154}        & \textbf{0.0294}        & \textbf{0.0519}        \\
\bottomrule
\end{tabular}
\end{adjustbox}
\end{table}

\subsection{GEP Performance under Distribution Shifts}\label{sec:gep_dist}
Results are shown in Table \ref{table:clean_data}. We make the following observations. Across all target datasets, and corresponding levels of distribution shifts, \gde~is by far the most effective at predicting generalization. For example, on the challenging STL10 benchmark, \gde~achieves improvements of $10\%$ and $30\%$ over \conf~and \lms~scores respectively. Similarly, \gde~provides consistent gains at all corruption levels on CIFAR-10-C. Notably, while \lms~was originally proposed in the context of sample-level scoring, it trails behind
\begin{wrapfigure}{r}{0.5\columnwidth}
\vspace{-0.1in}
\begin{center}
\includegraphics[width=0.5\columnwidth]{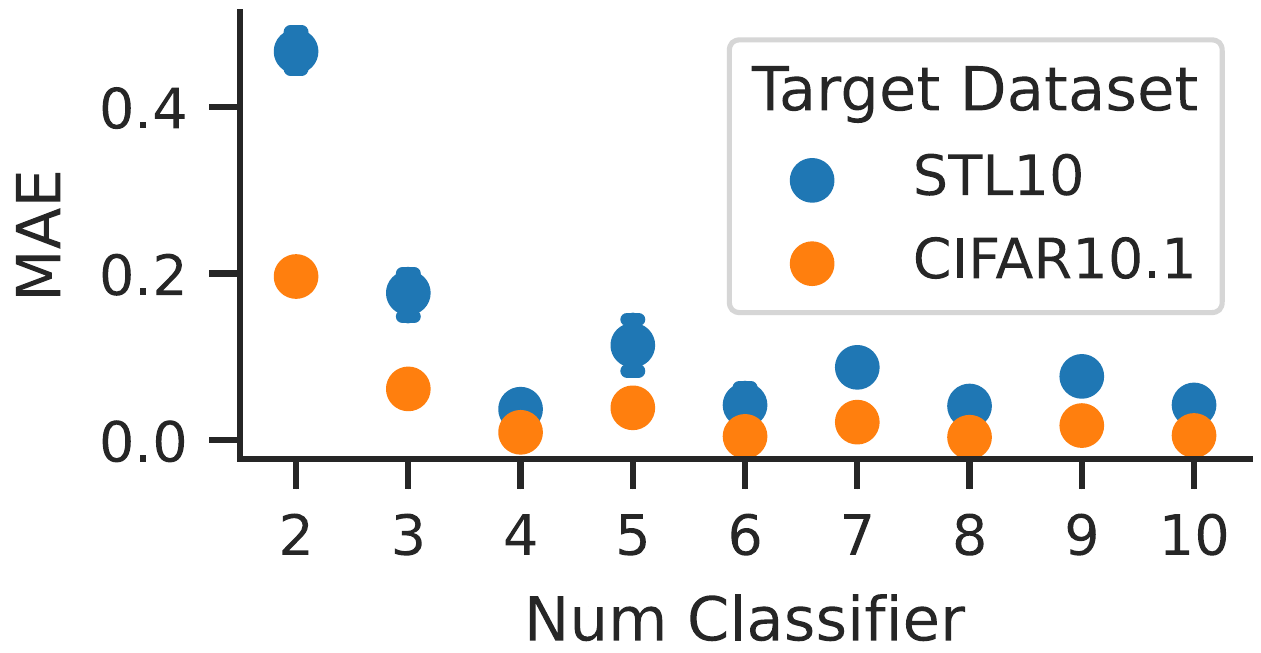}
\end{center}
\vspace{-0.2in}
\caption{\textbf{Effect of ensemble size on \gde~performance.}
\vspace{-0.1in}
}
\label{fig:num_classifier}
\end{wrapfigure}\conf, which was shown to more effective as a distribution-level scoring function by \cite{Guillory21_DoC}. While we used RandAug as it was shown to be effective by \cite{Ng22_LocalManifoldSmoothness}, our results here indicate that smoothness to such perturbations can fail to capture properties relevant for generalization under the real-world corruptions and shifts. This suggests that the choice of augmentation strategy is critical to \lms's effectiveness, and it is non-trivial to identify such a strategy without access to \textit{o.o.d} data.

Given that \gde~requires multiple independently trained models, we evaluate the effect of ensemble size in generating reliable scores. As shown in Fig. \ref{fig:num_classifier}, we find that the performance of \gde~begins to saturate at 4 models, though further increase in the ensemble size reduces the variability. In Sec. \ref{sec:corrupted_training}, we will show that the diversity
of the ensemble also plays an important role on GEP performance, particularly under challenging training conditions.

\begin{table}[t]
\centering
\caption{\small{\textbf{Effect of Data Fidelity Issues on GEP Performance}. We evaluate the effectiveness of scoring functions when the training data contains measurement error (MN), sampling discrepancies (US) and label noise (LN). Note, we include a variant of \gde~, where we improve the diversity of the ensemble by synthetically injecting label corruptions to $2\%$ of the samples.}}\label{tab:data_fidelity}
\begin{adjustbox}{max width=\columnwidth}
\begin{tabular}{lr|ccccc}
\toprule
\multirow{2}*{Dataset} & \multirow{2}*{$\mathrm{S}(\mathrm{x}; \mathrm{F})$} & \multicolumn{3}{c}{True Accuracy}    & \multicolumn{2}{c}{GEP Performance} \\
    \cmidrule(lr){3-5} \cmidrule(lr){6-7}
 &  &
  \multicolumn{1}{l}{CIFAR10} &
  \multicolumn{1}{l}{STL10} &
  \multicolumn{1}{l}{CIFAR10.1} &
  \multicolumn{1}{l}{STL10} &
  \multicolumn{1}{l}{CIFAR10.1} \\
 \midrule
\multirow{4}{*}{CIFAR10} &
  \conf &
  \multirow{2}{*}{0.9008} &
  \multirow{2}{*}{0.5484} &
  \multirow{2}{*}{0.8056} &
  0.1488 &
  {\ul 0.0117} \\
 & \lms         &                 &                 &                 & 0.3438            & 0.0961           \\
 & \gde         & \textbf{0.9408} & \textbf{0.5990} & \textbf{0.8665} & {\ul 0.0420}      & \textbf{0.0054}           \\
 & \gde$_{0.02}$ & {\ul 0.9334}    & {\ul 0.5953}    & {\ul 0.8555}    & \textbf{0.0115}   & 0.0179 \\
\midrule
\multirow{4}{*}{CIFAR10 (MN)} &
  \conf &
  \multirow{2}{*}{0.8284} &
  \multirow{2}{*}{0.5411} &
  \multirow{2}{*}{0.7155} &
  0.1010 &
  0.0206 \\
 & \lms         &                 &                 &                 & 0.3028            & 0.1464           \\
 & \gde         & \textbf{0.8782} & {\ul 0.5987}    & \textbf{0.7755} & {\ul 0.0528}      & \textbf{0.0108}  \\
 & \gde$_{0.02}$ & {\ul 0.8716}    & \textbf{0.6092} & {\ul 0.7670}    & \textbf{0.0219}   & {\ul 0.0123}     \\
 \midrule
\multirow{4}{*}{CIFAR10 (US)} &
  \conf &
  \multirow{2}{*}{0.9015} &
  \multirow{2}{*}{0.5943} &
  \multirow{2}{*}{0.7987} &
  0.1329 &
  0.0207 \\
 & \lms         &                 &                 &                 & 0.3300            & 0.1061           \\
 & \gde         & \textbf{0.9300} & \textbf{0.6439} & \textbf{0.8535} & {\ul 0.0451}      & {\ul 0.0130}     \\
 & \gde$_{0.02}$ & {\ul 0.9289}    & {\ul 0.6250}    & {\ul 0.8360}    & \textbf{0.0236}   & \textbf{0.0060}  \\
 \midrule
\multirow{4}{*}{CIFAR10 (LN)} &
  \conf &
  \multirow{2}{*}{0.8497} &
  \multirow{2}{*}{\ul 0.4736} &
  \multirow{2}{*}{0.7373} &
  {0.1802} &
  {\ul 0.0253} \\
 & \lms         &                 &                 &                 & 0.3569            & 0.1190           \\
 & \gde         & \textbf{0.9209} & { \textbf{0.5394}}    & \textbf{0.8315} & {\ul 0.0709}            & 0.0307           \\
 & \gde$_{0.02}$ & {\ul 0.9105}    & 0.4611          & {\ul 0.8060}    & \textbf{0.0601}   & \textbf{0.0153}\\ 
 \bottomrule
\end{tabular}
\end{adjustbox}
\end{table}

\subsection{Impact of Training Data Fidelity on GEP}\label{sec:corrupted_training}
In the preceding section, we evaluated the effectiveness of different scoring functions on a large family of image-level corruptions and distribution shifts. While we found \gde~and \conf~to be particularly effective, here, we seek to further evaluate their efficacy in the realistic, but more challenging setting where \textit{training data} may be compromised. Specifically, we consider \textit{label noise, measurement errors and sampling discrepancies} as sources of low-fidelity training data. 
We focus on these particular sources as they not only represent situations that are likely to be encountered in practical scenarios, but they also compromise the training data at different granularities, namely sample-, dataset-, and distribution- level. 
Given that GEPs are used as failure indicators, it is critical to evaluate the scoring functions in such settings. Additionally, we evaluate a variant of \gde~designed to increase ensemble diversity and further improve its performance. 

\noindent \textit{Experimental Setup}. The following processes are used to create compromised data: (i) \textbf{Label Noise:} We randomly select 5\% of the training set and randomly flip their labels. (ii) \textbf{Measurement Noise:} We first apply a Gaussian blur ($\sigma_1 = 0.5$) and then add standard normal noise ($\sigma_2 = 0.07$) to all training images; (iii) \textbf{Under-Sampling:} 20\% of the samples are randomly dropped from the \textit{automobile}, and \textit{bird} classes. Given these compromised datasets, we follow the same experimental setup introduced in Sec. \ref{sec:method}. However, models are now trained for 250 epochs, instead of 200 epochs, to achieve acceptable convergence. In the following analysis, we specifically focus on the near (CIFAR10.1) and far (STL10) distribution shift settings respectively.

\noindent \textbf{\gde~with improved diversity}. Motivated by the saturating effect of ensemble size in Fig. \ref{fig:num_classifier}, we propose a simple variant to \gde~that is designed to improve the diversity of the ensemble by synthetically corrupting the data using low levels of label noise. Here, the intuition is that such label noise requires models to learn slightly different functions to effectively minimize the loss on the mis-labeled samples. Since these randomly-labeled points account for a small portion (2\%) of the overall dataset, they generally do not substantially affect the ensemble accuracy (see Table \ref{tab:data_fidelity}). We denote this variant, \gde$_{0.02}$, and discuss its behavior, in addition to other scoring functions below (see Table \ref{tab:data_fidelity}).

\noindent {\ul Obs. 1}: \gde~and \conf~remain considerably more effective than \lms~even with compromised data. This is to be expected as the considered data infidelities are not expected to change the type of smoothness that is indicative of generalization on STL10 and CIFAR10.1.
\\
\noindent {\ul Obs. 2}: While adding measurement noise (MN) and under-sampling (US) does minimally harm the true target accuracy, we see GEP performance of \lms~and \conf~improves on STL10--even better than their performance obtained by training on the clean dataset. We posit in Sec. \ref{sec:simp_bias} that this can be attributed to decreased reliance upon simple features that lead to over-confident, but ultimately misleading, scores.
\\
\noindent {\ul Obs. 3}: In contrast, on the near \textit{o.o.d} setting of CIFAR10.1, we not only see that the true target accuracy decreases, but also that GEP performance decreases across all methods relative to training on the clean dataset. Given the closeness of CIFAR10.1 to CIFAR10, it is likely that low fidelity data struggles to capture the features necessary for generalization or estimation on the target distribution. 
\\
\noindent {\ul Obs. 4}: Incurring only a small drop in the true target accuracy, our variant \gde$_{0.02}$ improves GEP performance on STL (over 50\% on CIFAR10, CIFAR10 (MN), CIFAR10 (US)), and maintains comparable GEP performance on the near \textit{o.o.d} setting. While target accuracy does decrease noticeably on CIFAR10 (LN), we note that this is an edge case, as the underlying dataset has already been perturbed by label noise. The efficacy of \gde$_{0.02}$ on CIFAR10.1 is surprising, as the perturbed distributions are certainly further from target than the clean source distribution, but this suggests diversity plays a critical role in obtaining effective scores. 

\begin{table}[t]
\centering
\caption{\small{\textbf{Effect of Simplicity Bias on GEP Performance}. We evaluate the effectiveness of GEPs when the training data and target data contain varying degrees of correlation between simple/complex features.  MAE is reported, and we \textbf{bold} the best score per correlation level. Observe that scores suffer when the correlation is broken.}}
\begin{adjustbox}{max width=\columnwidth}
\begin{tabular}{lcccccc}
\toprule
\multirow{2}{*}{$\mathrm{S}(\mathrm{x}; \mathrm{F})$} & \multicolumn{2}{c}{Corr. = 0.95}                       & \multicolumn{2}{c}{Corr. = 0.99}                       & \multicolumn{2}{c}{Corr. = 1.0}                        \\
 \cmidrule(lr){2-3} \cmidrule(lr){4-5} \cmidrule(lr){6-7}
 & \multicolumn{1}{c}{Corr.} & \multicolumn{1}{c}{Rand.} & \multicolumn{1}{c}{Corr.} & \multicolumn{1}{c}{Rand.} & \multicolumn{1}{c}{Corr.} & \multicolumn{1}{c}{Rand.} \\
 \midrule
\conf & 0.0184 & \textbf{0.0350} & 0.0136 & \textbf{0.2489} & \textbf{0.0005} & \textbf{0.7927} \\
\lms & 0.0557 & 0.2438 & \textbf{0.0056} & 0.5571 & 0.0028 & 0.8699 \\
\gde & \textbf{0.0084} & 0.1262 & 0.1019 & 0.5125 & \textbf{0.0007} & 0.8492 \\
\bottomrule
\end{tabular}
\label{tab:simp_bias}
\end{adjustbox}
\end{table}
\section{Effect of Simplicity Bias on GEPs}
\begin{figure}
    \centering
    \includegraphics[width = 0.75\columnwidth]{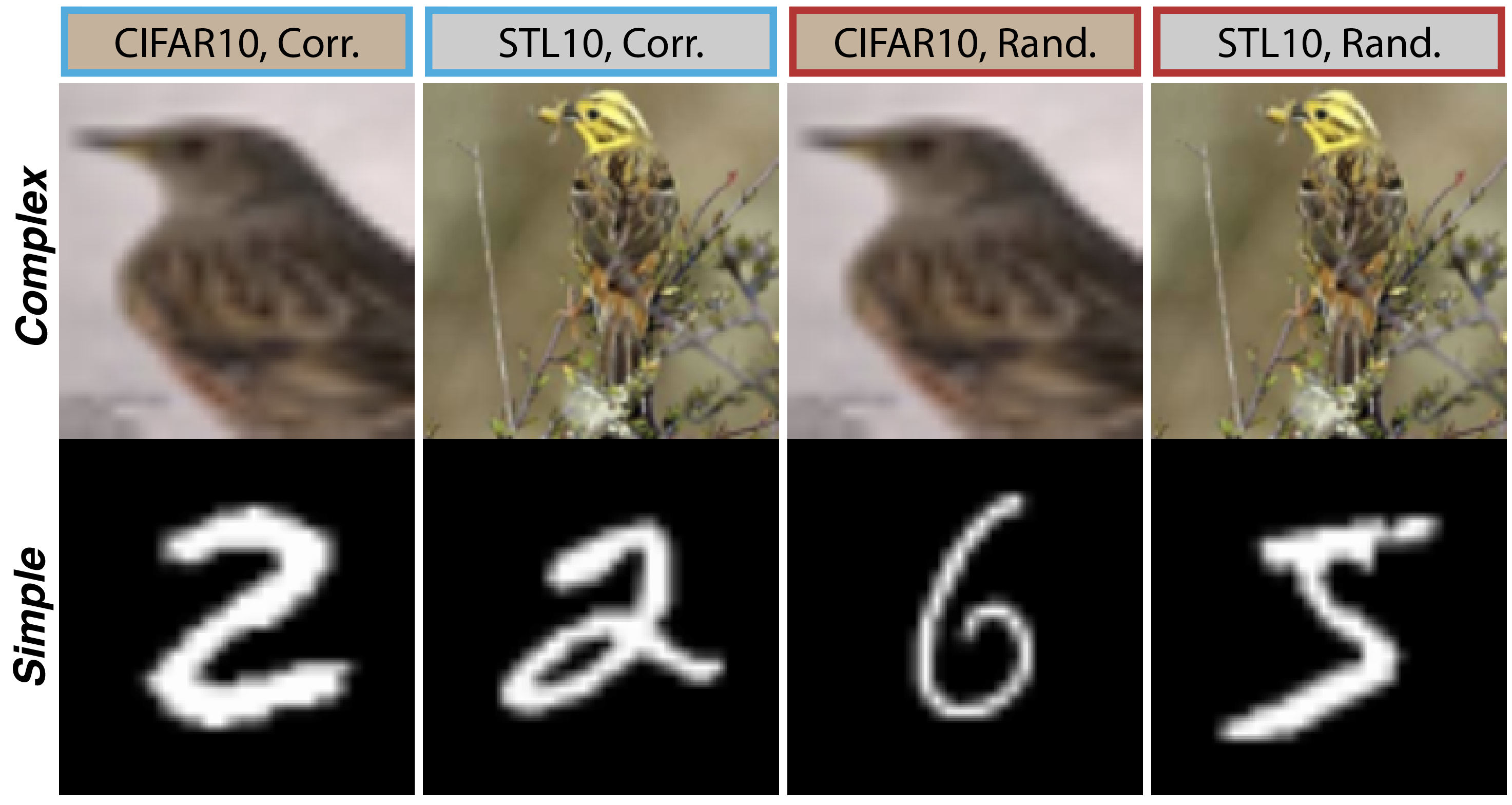}
    \caption{\textbf{Simplicity Bias Dataset, (Fig. 2 \cite{Trivedi22_Adaptation}).} Dominoes comprised of complex (CIFAR10) and simple (MNIST) features are used to control the simplicity bias on target datasets.}
    \label{fig:cifar_mnist}
    \vspace{-0.3cm}
\end{figure}

In the preceding section, we found that training on compromised data may in fact decrease the GEP error. We hypothesize that this decrease can be partially attributed to the noisy datasets, mitigating \textit{simplicity bias}, i.e., the well known propensity of deep neural networks to rely upon simple, spurious features in lieu of more complex/expressive ones~\cite{brutzkus18_sgd,Gunasekar18,Shah20_SimplicityBias,geirhos18_texturebias}. Given that simple features are not expected to generalize on \textit{o.o.d} datasets, but DNNs remain susceptible to relying upon such simple features, we posit that GEPs will also see decreased performance on distributions where simple features are no longer indicative of the label. We test this hypothesis using a synthetic setting that controls the discriminability of simple features on target datasets, as discussed below.
\\
\noindent \textit{Experimental Setup.} We use a custom ``dominoes" dataset \cite{Shah20_SimplicityBias} of complex and simple features by pairing each class from CIFAR10 (complex feature) with the corresponding digit class in MNIST (simple feature) \cite{Trivedi22_Adaptation}. (See Fig. \ref{fig:cifar_mnist}). Three levels of correlation (95\%, 99\%, 100\%) between the target and simples features are considered during training.  When predicting generalization, we sample complex features from STL10, as well as create a variant that randomizes the spurious correlation between simple and complex features. We fine-tune a MoCo-V2 pretrained ResNet-50~\cite{He20_MoCo} for 20 epochs with lr=0.001 and average results over $3$ seeds.

As shown in Table. \ref{tab:simp_bias}, we see that the prediction error often substantially increases when evaluating on the randomized target dataset, e.g., where the simple feature is no longer predictive. While we would expect that the target accuracy decreases, the decreased GEP performance is particularly troubling as such methods are intended to detect these very failures. Moreover, we note that as the correlation between the simple and complex feature increases (Corr=0.95 vs. Corr=1.0), the gap between GEP's performance on the Corr. and Rand. variants of the target dataset increases. Indeed, the Corr. MAE decreases as the training dataset correlation increases (Corr=0.95 vs. 1.0), but the Rand. MAE increases. This result further highlights the harmful role of simplicity bias on GEP performance. 
\label{sec:simp_bias}
\vspace{-0.2pt}
\section{Conclusion}
\label{sec:concl}
In this work, we rigorously studied the design of scoring functions in GEPs and found that their choice is critical to produce consistently reliable predictors across different distribution shifts and noise corruptions. In fact, when the GEP construction does not involve calibration datasets or training a large family of models, even state-of-the-art scoring functions such as \conf~and \lms~can struggle. In comparison, we found \gde~to be a more reliable alternative. 
Furthermore, using our new \gde~variant, we demonstrated that improving diversity of the ensemble leads to well-calibrated GEPs, while incurring only a small drop in target accuracies. Finally, in a controlled empirical setting, we showed how reliance on simple features can adversely affect the GEP performance. 



\bibliographystyle{ieee}
\bibliography{icassp}

\end{document}